\documentclass[sigconf]{acmart}

\usepackage{pifont}
\usepackage{amsmath}
\usepackage{graphicx}
	
\usepackage{color, colortbl}

\definecolor{LightCyan}{rgb}{0.88,1,1}
\definecolor{Gray}{gray}{0.9}
\definecolor{LightOrange}{RGB}{255,204,153}
\definecolor{LightPink}{RGB}{255,182,193}
\definecolor{LightGreen}{RGB}{144,238,144}
\definecolor{corn}{rgb}{0.98, 0.93, 0.36}

\AtBeginDocument{%
  }

\setcopyright{rightsretained}
\copyrightyear{2024}
\acmYear{2024}
\acmDOI{00.000}

\acmConference[]{}{}{}

\acmISBN{--}

\begin{document}

\title{CLIP-VAD: Exploiting Vision-Language Models for \\ Voice Activity Detection}

\author{Andrea Appiani}
\affiliation{%
\department{Dep. of Management, Information and Production Engineering}
  \institution{University of Bergamo}
  \city{Bergamo}
  \country{Italy}
}
\email{appiani.andrea@gmail.com}

\author{Cigdem Beyan}
\affiliation{%
  \department{Department of Computer Science}
  \institution{University of Verona}
  \city{Verona}
  \country{Italy}
}
\email{cigdem.beyan@univr.it}

\renewcommand{\shortauthors}{A. Appiani \& C. Beyan}

\begin{abstract}

Voice Activity Detection (VAD) is the process of automatically determining whether a person is speaking and identifying the timing of their speech in an audiovisual data.
Traditionally, this task has been tackled by processing either audio signals or visual data, or by combining both modalities through fusion or joint learning. In our study, drawing inspiration from recent advancements in visual-language models, we introduce a novel approach leveraging Contrastive Language-Image Pretraining (CLIP) models. The CLIP visual encoder analyzes video segments composed of the upper body of an individual, while the text encoder handles textual descriptions automatically generated through prompt engineering. Subsequently, embeddings from these encoders are fused through a deep neural network to perform VAD. Our experimental analysis across three VAD benchmarks showcases the superior performance of our method compared to existing visual VAD approaches. Notably, our approach outperforms several audio-visual methods despite its simplicity, and without requiring pre-training on extensive audio-visual datasets.

\end{abstract}

\keywords{Active speaker, voice activity detection, social interactions, vision language models, panel discussions}

\maketitle

\section{Introduction}
\label{sec:intro}

Voice Activity Detection (VAD) is the process of automatically determining whether a person is speaking or not in a recording, thereby addressing the question of ``Who is speaking and When?''. This task is crucial in various real-world applications such as human-robot interaction \cite{skantze2021turn}, speech diarization \cite{xu2022ava,wang2018speaker,chung2020spot}, multiparty dialogues among humans\cite{hung2009speech}, social behavior analysis \cite{beyan2019sequential}, automatic speech recognition \cite{gorriz2010improved}, speech enhancement \cite{michelsanti2021overview}, and emotion recognition \cite{moine2021speaker}.

VAD has traditionally been approached through audio processing \cite{moattar2009simple,minotto2014simultaneous,patrona2016visual}, which presents challenges, particularly in scenarios like unstructured social gatherings where multiple speakers talk simultaneously or when there is a high number of subjects or in case of a close proximity between speakers. With the advancement of Convolutional Neural Networks (CNNs), the integration of video modality alongside audio has markedly enhanced VAD performance \cite{tao2021someone,kopuklu2021design}. 
However, audio-visual VAD still faces challenges, particularly in effectively modeling both modalities jointly or requiring speech enhancements as shown in \cite{xiong2022look}. Furthermore, audio-visual VAD may not always be applicable due to technical, ethical, or legal considerations regarding audio \cite{VAD_Motion_Segmentation,RealVAD}. Consequently, some approaches emphasize the importance of addressing this task solely based on visual data. Indeed, in some instances, visual VAD has demonstrated superior performance compared to audio-visual VAD \cite{VAD_UBM,VAD_Motion_Segmentation,RealVAD}, highlighting the crucial role of the video modality in this domain.

The utilization of head crops, especially the regions encompassing individuals' faces, stands out as the primary visual cues employed in both visual VAD and audio-visual VAD \cite{zhang2021unicon,xiong2022look,tao2021someone,kopuklu2021design,tao2021someone}.
While some studies have concentrated on linking lip movements with audio signals \cite{chung2018learning,liu2011visual,sodoyer2009study,chung2017out}, others have utilized facial landmarks \cite{9931697,huang2020improved}. Despite the latest advancements in audio-visual VAD often relying on facial images to achieve state-of-the-art (SOTA) results, some research emphasizes the importance of analyzing upper body motion \cite{VAD_Motion_Segmentation,RealVAD,VAD_UBM} and gestures \cite{cristani2012look,gebre2013gesturer}, proving the relevance of these cues in detecting voice activity. 

Our proposal follows the findings of \cite{VAD_UBM,VAD_Motion_Segmentation,RealVAD,primitive}, emphasizing the significance of processing \textbf{upper body images} for VAD. It most closely intersects with visual VAD since the input to our model is a video segment that is associated with the VAD label of an individual as \textit{speaking} or \textit{not speaking}. On the other hand, our initial tests with \textbf{Vision-Language Models (VLMs)}, especially when provided with upper body crop images, along with insights from studies like \cite{xenos2024vllms} that focus on emotion recognition, reveal the proficiency of VLMs in describing images by focusing on facial muscles and gestures, which is crucial for capturing essential cues for VAD. 
Encouraged by these findings, we opt to leverage the \textbf{text descriptions} automatically generated by VLMs using prompt engineering when the input image represents the central frame of a video segment depicting an individual. Additionally, motivated by CLIP's \cite{CLIP} enhanced performance across various visual downstream tasks \cite{CLIP_depth,CLIP_emotion,CLIP_gaze_prediction}, we adopt pre-trained CLIP models. These models employ separate encoders for visual and textual data, which are aligned within a shared embedding space using contrastive loss \cite{CLIP}. 

In our network, referred to as \textbf{CLIP-VAD}, the visual encoder processes video segments containing upper body images, while the text encoder handles textual descriptions provided by a VLM. These encoders produce embeddings, which are then concatenated and fused using a deep neural network to perform the VAD task. While our approach is not inherently multimodal in terms of data since focusing on video segments comprising only upper body frames, it is crucial to note the VLM's ability to interpret not just body movements like arm gestures, but also facial activity. On the other hand, our proposed architecture is multimodal, incorporating both visual and textual cues. To our knowledge, this is the first time text data is being used for VAD.

CLIP-VAD, tested on three VAD benchmarks, demonstrates superior performance compared to all SOTA visual VAD methods. 
It also yields promising results, achieving performance levels comparable to or better than several SOTA audio-visual VAD approaches, despite not being pre-trained on large audio-visual VAD datasets. This underscores the effectiveness of our model, trained and tested directly on visual VAD benchmarks.

The main contributions and findings of our study can be summarized as follows: \\
\noindent (1) We introduce a novel VAD method that effectively utilizes visual-language pre-training techniques. To the best of our knowledge, this is the first work to adopt CLIP \cite{CLIP} for VAD. Consequently, we demonstrate that our model, CLIP-VAD, surpasses SOTA visual VAD methods, affirming the utility of joint learning of text descriptions and visual features. \\
\noindent (2) This is also the first attempt to employ VLMs with prompt engineering to perform VAD and to generate text descriptions corresponding to individuals' speaking activity when their upper body images are the inputs. While the standalone VLM model may not match the effectiveness of our CLIP-VAD for the VAD task, its text descriptions enhance the utilization of spatio-temporal upper body features, thereby improving CLIP-VAD's performance. \\
\noindent (3) Through extensive experimentation, we demonstrate that our approach outperforms all visual methods as well as a standalone VLM. Moreover, our CLIP-VAD consistently achieves results comparable to or surpassing the audio-visual SOTA, even if it employs a simpler pipeline compared to them and without the necessity of pre-training on audio-visual data.

The remainder of this paper is structured as follows. In Sec. \ref{sec:relWork}, we provide an overview of the existing literature on VAD and previous studies on CLIP, emphasizing the unique aspects of our approach. The proposed method, CLIP-VAD, is described in Sec. \ref{sec:relWork} together with its implementation details. Following that, in Sec. \ref{sec:exp}, we introduce the datasets, and the evaluation metrics we used in line with the SOTA, and also present a comprehensive ablation study. This section further includes a comparative analysis between CLIP-VAD and SOTA. Finally, Sec. \ref{sec:conc} summarizes the major findings of this study, highlights the limitations, and presents possible future directions.

\section{Related Work}
\label{sec:relWork}
In this section, we provide an overview of the current body of literature on Voice Activity Detection (VAD) and summarize the various applications of Contrastive Language-Image Pretraining (CLIP). Additionally, we outline the rationale for employing CLIP \cite{CLIP} and Vision-Language Models (VLMs) in the context of VAD.

\subsection{Voice Activity Detection}
\label{sec:VAD_RW}
Earlier works have tackled the task of VAD solely through audio signal processing, as evident from a broad literature, e.g., \cite{moattar2009simple,minotto2014simultaneous,patrona2016visual}. However, performing audio-based VAD can be challenging, especially in real-life scenarios where sounds may originate from multiple speakers simultaneously and when the speakers are nearby. 

Particularly, with the advancement of Convolutional Neural Networks (CNNs), visual information has also been integrated into audio-based VAD, resulting in the development of several multimodal VAD approaches that employ both audio and visual cues. These studies, which perform audio-visual VAD (also referred to as audio-visual active speaker detection), typically consider the temporal dependency between audio and visual data. This involves the application of Recurrent Neural Network (RNN) \cite{tao2017bimodal,tao2019end}, Gated Recurrent Unit (GRU) \cite{roth2020ava}, Long Short-Term Memory (LSTM) \cite{sharma2020crossmodal,shvets2019leveraging}, and Transformer Layer \cite{tao2021someone}. Audio-visual VAD, when fully leveraging cross-modal synchronization information, can achieve highly successful performance \cite{zhang2021unicon,xiong2022look}. However, much of the existing work relies on separately encoding the unimodal features of audio and video, limiting the exploitation of cross-modal synchronization information. For instance, in extracting visual features, some studies employ 3D CNNs to capture temporal dependencies from video data \cite{tao2021someone,kopuklu2021design}. Conversely, for audio features, CNNs are often utilized with log-Mel or Short-Time Fourier Transform (STFT) spectrograms as inputs \cite{tao2021someone}, or directly applied to the audio waveform \cite{kopuklu2021design}. Furthermore, in many studies, visual information is primarily employed to link the active speaker with speech, reflecting a methodology that does not fully leverage visual data \cite{gebru2017audio}. It is worth noting that audio-visual VAD techniques utilize the head crops of individuals to extract visual features, sometimes with a focus on lip movements \cite{VAD_Audio_Supervision,chung2017out}. However, such approaches are particularly effective only for frontal facial images.

On the other hand, numerous methods exclusively rely on visual data to address VAD. These visual VAD approaches can be categorized into two groups: those that analyze facial cues, such as facial landmarks \cite{chung2018learning,joosten2015voice,haider2016active,stefanov2017vision,hung2009speech,stefanov2019self}, and those that consider body cues, including hand gestures, head movements, and upper body motion \cite{cristani2012look,gebre2013gesturer,VAD_Motion_Segmentation,RealVAD,VAD_UBM}. Interestingly, some visual VAD studies, despite lacking audio, have demonstrated superior performance compared to a few audio-visual methods \cite{VAD_Motion_Segmentation,RealVAD,VAD_UBM}. Overall, visual VAD studies offer a significant alternative, especially in scenarios where accessing the audio signal is not feasible due to functional, legislative, or moral limitations \cite{stefanov2017vision,VAD_Audio_Supervision,VAD_Motion_Segmentation,RealVAD,VAD_UBM}.

In this study, inspired by the improvements in various visual recognition tasks achieved through Vision-Language Models (VLMs), we introduce a method that combines visual and text modalities to tackle the VAD task. To do so, we particularly utilize CLIP \cite{CLIP} while the text input is produced through prompt engineering. To the best of our knowledge, this is the first work using video clips and associated generated text within a CLIP framework \cite{CLIP} to perform VAD. The proposed architecture is much simpler for example compared to \cite{xiong2022look} simultaneously applying speech enhancement, or \cite{tao2021someone} performing long-term
temporal intra-speaker context processing. Besides, we do not require using the pre-trained weights
on large-scale VAD datasets as applied in \cite{zhang2021unicon}.

\begin{figure*}[ht!]
\centering
\setlength{\abovecaptionskip}{5pt}
    \includegraphics[width=0.95\linewidth]{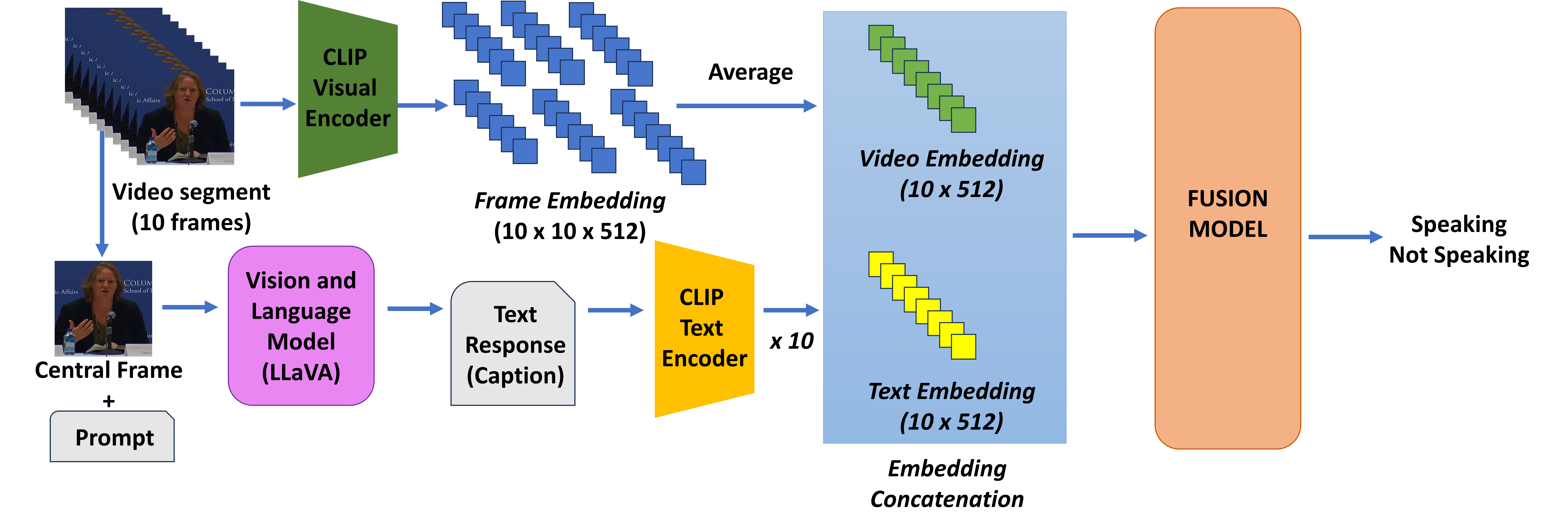}
    \caption{The overview of the proposed approach: CLIP-VAD. 
    Our approach entails short video segments capturing individuals' upper body, along with textual descriptions regarding their speaking status, derived through prompt engineering. The goal is to harness both video and text embeddings to enable a fusion network to determine whether the person depicted is speaking or not. 
    The input video segment consists of 10 frames, each represented as an embedding of size 10$\times$512 after being input to the CLIP visual encoder. These 10-frame embeddings, which are 10 $\times$ 10 $\times$ 512, are averaged along the temporal channel. 
    The central frame of these 10 frames, together with a prompt, is input to the LLaVa model \cite{Llava_paper1, Llava_paper2} to generate a textual response (caption). The caption is then provided to the CLIP text encoder, resulting in a single text embedding. This text embedding is replicated 10 times and concatenated with the 10$\times$512 video embeddings to be given as an input to a Fusion Model designed as either an MLP or a Transformer network to predict the VAD label.}
    \label{fig:proposed}
    \Description{Overview of the proposed method.} 
\end{figure*}

\begin{figure}[h!]
\centering
\setlength{\abovecaptionskip}{5pt}
    \includegraphics[width=1\linewidth]{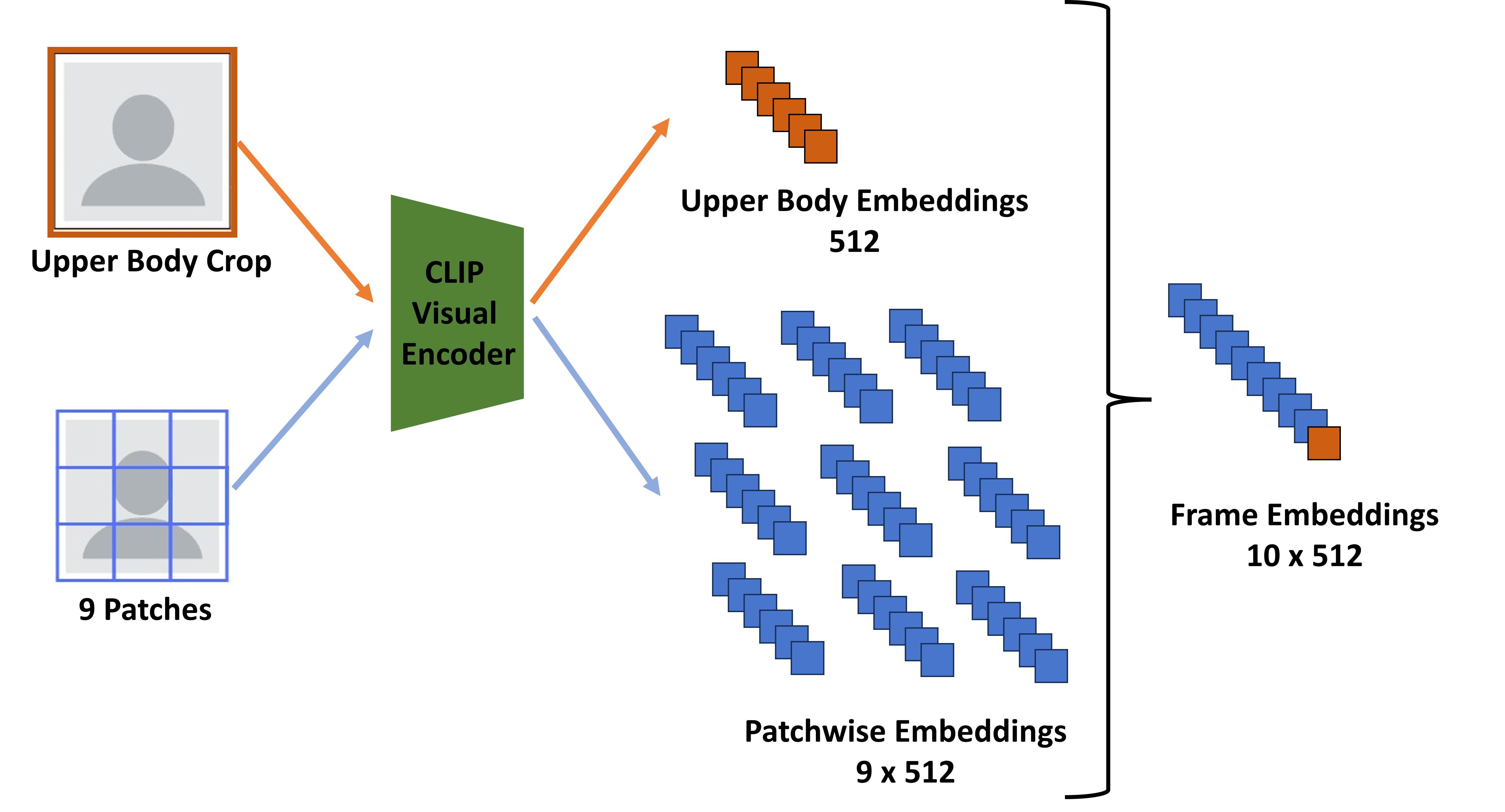}
    \caption{The extraction of visual embeddings involves using the CLIP visual encoder. That encoder takes a frame consisting of an upper body crop of an individual along with the 9 non-overlapping patches obtained from that frame. This process captures both local and global features. From 10 inputs, we obtain a frame embedding size of 10$\times$512 (1 $\times$ upper body embeddings + 9 $\times$ patchwise embeddings). This procedure is repeated for all frames within a given video segment.}
    \label{fig:patches}
    \Description{Extraction of visual embeddings.} 
\end{figure}

\subsection{CLIP}
\label{sec:clip}
A VLM processes images and their corresponding textual descriptions, learning to connect information from both modalities. The visual component of the model usually captures spatial features from images, while the language model encodes information from text. CLIP, which stands for Contrastive Language-Image Pretraining, is one of the most prominent models in this category. 
The usage of CLIP includes zero-shot classification and fine-tuning for downstream tasks, as demonstrated in various papers such as \cite{CLIP,CLIP_finetune,CLIP_aiuto_VL}. 
Unlike frequently applied downstream tasks such as image classification \cite{CLIP_patches,srivastava2023retailklip} and image segmentation \cite{CLIP_driven_segmentation, CLIP_organs}, CLIP is also utilized in diverse applications such as image enhancement \cite{CLIP_controluce}, monocular depth estimation \cite{CLIP_depth}, text-to-shape generation \cite{CLIP_textTOshape}, image manipulation \cite{CLIP_manipulation}, medical image processing \cite{CLIP_organs}, anomaly detection \cite{CLIP_out_distribution}, gaze estimation \cite{CLIP_gaze_prediction}, emotion recognition \cite{CLIP_emotion,wang2024emoasst,garg2022video} and so forth. 

Recent studies, such as \cite{CLIP_emotion}, emphasize CLIP's ability to extract robust features for representing facial images and expressions, both through fine-tuning and zero-shot learning. 
Furthermore, our investigations with other VLMs, particularly LLaVA-13B \cite{Llava_paper1, Llava_paper2}, have highlighted the potential of these models (see Sec. \ref{sec:exp} for results) in describing a speaking individual. We observed that LLaVA-13B demonstrates adeptness in focusing on hand gestures, facial expressions, and the shape of the mouth. 
Additionally, the research presented in \cite{xenos2024vllms} provides compelling evidence for the efficacy of LLaVA in facial emotion recognition.
All these findings have strengthened our confidence in utilizing LLaVA for generating text descriptions through prompt engineering in conjunction with using CLIP to realize VAD.

\section{Proposed Method: CLIP-VAD}
\label{sec:method}
The structure of CLIP-VAD is depicted in Fig. \ref{fig:proposed}. It consists of both a visual and a textual component, predominantly leveraging the CLIP architecture \cite{CLIP}. Our methodology involves short video segments capturing individuals' upper body, accompanied by textual descriptions regarding their speaking status, which are extracted through prompt engineering. 
The aim is to harness both video embeddings capturing temporal information and text embeddings to strengthen a neural network in determining whether the depicted individual is speaking or not.


\subsection{Preliminaries}
\label{sec:Clip}
Contrastive Language-Image Pre-training (CLIP) \cite{CLIP} employs a dual-encoder framework consisting of a visual encoder denoted as $\mathbf{E_{v}}$ and a text encoder termed as $\mathbf{E_{t}}$. $\mathbf{E_{v}}$ processes input images $I \in \mathbb{R}^{H \times W \times 3}$ by dividing them into a sequence of fixed-size patches. These patches, combined with a learnable class token, are transformed to a unified vision-language embedding space, resulting in a final visual feature $f_v = \mathbf{E_{v}}(I) \in \mathbb{R}^d$ when $d$ represents the dimensionality of the features. On the other hand, $\mathbf{E_{t}}$ converts textual input (e.g., prompt) shown as $Tx$ into text embeddings, augmenting them with a learnable class token to create an input feature matrix. This matrix is then processed to extract textual features $f_{t} = \mathbf{E_{t}}(Tx)$. By aiming to maximize the similarity denoted as $\mathcal{\texttt{sim}}$ between the matched text-image pairs and to minimize the $\mathcal{\texttt{sim}}$ for the unmatched pairs, CLIP is trained using a contrastive loss function. CLIP uses textual prompts to generate specific text features and compute the prediction by calculating the distance to an image feature as:
\begin{equation}
\label{eq:primary}
\mathcal{P}(y|v) = \frac{\mathrm{exp}(\texttt{sim}(f_v, f_t^y)/\tau)}{\sum_{i=1}^{K} \mathrm{exp}(\texttt{sim}(f_v, f_t^i)/\tau)}
\end{equation}
where $\tau$ is the temperature parameter, controlling the scale or smoothness of the probability distribution.

\subsection{Formal description}
\label{sec:formalDesc}

Given a video segment $\mathbf{V = \{v_1, v_2, \ldots, v_T\}}$ consisting of $T$ frames and the \textbf{VAD label} $L$, we initially preprocess these inputs to enhance the utilization of the pre-trained CLIP model. Therefore, the input frames are resized to match the input size expected by \textbf{CLIP's visual encoder}, which can be either a Transformer or a Residual Network. These resized frames are then embedded into a set of visual tokens $T \times N \times D$, where $N$ is the number of tokens and $D$ is the dimension of each token. Subsequently, we compute the average of these embeddings to obtain a tensor $F_v \in \mathbb{R}^{N \times D} = \{f_{v_{1}}, f_{v_{2}}, \ldots, f_{v_{T}}\}$ along the temporal channel.

The central frame of the video segment, denoted as $\{v_{T/2}\}$, is utilized to generate text input for \textbf{CLIP's text encoder}. Prompt engineering is employed, where $\{v_{T/2}\}$ is paired with a prompt and provided to a \textbf{Vision Language Model (VLM)} to generate text responses. These responses are then passed to the text encoder of CLIP, resulting in text tokens $F_t \in \mathbb{R}^{N \times D} = \{f_{T/2}, \ldots,  f_{T/2}\}$ where $\{f_{T/2}\}$ is replicated to arrive to the size of $F_v$. 
Finally, both the visual and textual embeddings are concatenated [$F_v$, $F_t$] and fed as input to the \textbf{fusion model $FN$} for classifying the entire video segment as either speaking or not speaking.

The $\mathbf{FN}$ is designed as either a Multilayer Perceptron (MLP) denoted as $\mathbf{FN_{MLP}}$ or a Transformer referred to as $\mathbf{FN_{T}}$, noticing that depending on the size of the training data, one model may outperform the other. Empirical evidence in the next section demonstrates that $\mathbf{FN_{T}}$ typically requires more training data compared to $\mathbf{FN_{MLP}}$ to achieve better performance. $\mathbf{FN_{T}}$ includes a normalization layer to process the input, that is transformed into three branches, i.e., $Q$, $K$, and $V$.
Multi-head self-attention described in terms of self-attention layers are defined as $MLP(SA(Q,K)V)$, where $SA(Q,K) = \text{Softmax}(\frac{QK^{Tr}}{\sqrt{c}})$ when $c$ denotes feature dimension and $Tr$ is the transpose operation. The outputs are given to the classification head for VAD. On the other hand, $\mathbf{FN_{MLP}}$ consists of multiple dense layers, including an input layer, multiple hidden layers, and an output layer. Each neuron applies an activation function to the weighted sum of its inputs, introducing non-linearity into the network and enabling it to learn complex patterns in the data.

\subsection{Implementation Details}
\label{sec:ImpDet}
The input $V$ comprises frames showing the upper body of an individual for a fixed duration of 10 frames, each sharing the same ground-truth label, as described in \cite{VAD_UBM,VAD_Motion_Segmentation}. During training, if there is no video segment sharing the same ground-truth data for 10 frames, the remaining frames are repeated until a video segment of 10 frames is obtained.

We conducted experiments with pre-trained CLIP models, encompassing Residual Networks and Vision Transformers (ViT), which are ResNet101 and ViT-B/16, respectively. Each of these models includes both a visual and a text encoder, accompanied by a preprocessing function and a tokenizer. The preprocessing function is responsible for adapting images to the format accepted by the visual encoder, while the tokenizer divides the text into tokens suitable for input into the text encoder \cite{CLIP}.

Each frame within a $V$ undergoes resizing to dimensions of 224$\times$224 to match the input image size expected by the CLIP visual encoder. Subsequently, the frame is partitioned into 9 non-overlapping patches. Both these patches (allowing us to capture the local features) and the complete upper-body image (enabling us to extract global features) are then fed into a visual encoder, producing embeddings of size 512 for each. Consequently, the output for each frame yields a vector of dimensions 10$\times$512 (see Fig. \ref{fig:patches}). This process is repeated for every frame in the $V$, resulting in a total of 100 embeddings with the dimension of 512 (10$\times$10$\times$512). Finally, the average is computed along the temporal dimension to consolidate the embeddings into a final vector of dimensions 10$\times$512, representing the entire $V$. 

As a VLM, we selected LLaVA-13B \cite{Llava_paper1,Llava_paper2}. The frame $\{v_{T/2}\}$, in our case the 5th frame for a $V$ composed of 10 frames, was chosen and provided as an input to LLaVA-13B, along with a textual prompt, to generate a textual response. We experimented with two prompts: \textit{1) Is the person speaking? Answer with yes or no.} and \textit{2) Is the person speaking? Explain why in a few words}. The first prompt consistently yields a response as ``yes'' or ``no''. For the second prompt, we set the temperature and the maximum number of tokens to 0.2 and 50, respectively. 
While there is a single set of textual tokens obtained for each video segment, there are visual tokens for every frame within that video segment. To address this imbalance, our solution is to replicate the textual tokens for each frame, aligning the number of textual tokens with the number of visual tokens. This preserves the significance of textual information without any reduction in its importance.

The $\mathbf{FN_{T}}$ takes inputs consisting of 20 tokens: 10 visual and 10 textual. It includes two attention heads, two linear layers responsible for increasing the dimensionality of the embeddings from 512 to 768, and linear layers responsible for classification, along with two normalization layers processing the classification results to output the final logits. 
In $\mathbf{FN_{MLP}}$, there are four linear layers with input sizes of 1024, 512, 256, and 1, respectively, with the last layer corresponding to the classification layer. ReLU is used as the activation function, and each linear layer is followed by batch normalization. Similar to the $\mathbf{FN_{T}}$, $\mathbf{FN_{MLP}}$ also takes concatenated visual and textual embeddings as input, without considering tokens, resulting in a single embedding of size 1024.

The learning rate for both our $\mathbf{FN_{T}}$ and $\mathbf{FN_{MLP}}$ models was set to different values: 0.01, 0.001, and 0.0001, with a weight decay of $1e^{-4}$. We trained the models for up to 50 epochs using the Adam optimizer. During training, we utilized the Binary Cross-Entropy Loss with Logits (BCEWithLogLoss) as the loss function. The batch size, following the settings in \cite{VAD_UBM,RealVAD}, was set to 128, with 64 speaking and 64 not speaking randomly selected $V$ segments used in each batch. We also employed the same data augmentation procedure applied in \cite{VAD_UBM,RealVAD}.

\begin{table*}
    \centering
    \caption{The ablation study evaluates the performance of various combinations of CLIP-VAD on the Modified Columbia dataset \cite{VAD_Motion_Segmentation} in terms of the F1-score (\%). The best results are highlighted in black, while the second-best results are \underline{underlined}. Vis. stands for visual modality and $\mathbf{E_{v}}$ represents the visual encoder. Exp. indicates the index of the experiments. See text for the descriptions of \textit{prompt 1}, fixed and variable.}
    \label{tab:ablation}
    \begin{tabular}{llccccccccll}  \hline \hline
            \textbf{Exp.} & \textbf{Model}&  \textbf{Modality}&  \textbf{$\mathbf{E_{v}}$} &  \textbf{Text Prompt}&  \textbf{Bell}&  \textbf{Sick} &  \textbf{Long}&  \textbf{Boll.}& \textbf{Lie.} & \textbf{Avg.}&\textbf{Std.}\\ \hline \hline  
1 & LLaVA-13B \cite{Llava_paper1,Llava_paper2} &  Vis. \& Text &  X &  prompt 1 &   82.06 & 54.89 & 41.41 & 76.66 & 76.97 & 66.40          & 17.46 \\

2 & ViT-B/16 + MLP &  Vis. &  pre-trained & X & 66.53 & 73.88 & 44.94 & 83.6  & 94.96 & 72.78          & 18.87 \\
3 & ResNet101 + MLP &  Vis. &  pre-trained &  X & 66.77 & 84.99 & 78.76 & \underline{85.25} & 90.77 & 81.31          & 9.17  \\
4 & ResNet101 + MLP &  Vis. &  fine-tuned &  X  & 71.29 & 89.98 & 81.42 & 83.08 & 88.88 & 82.93          & 7.46  \\
5 & $\mathbf{FN_{MLP}}$ & Vis. \& Text & pre-trained & fixed & 75.31 & 91.59 & 81.61 & \textbf{86.73} & 83.95 & 83.84          & 6.04  \\
6 & $\mathbf{FN_{T}}$ & Vis. \& Text & pre-trained & fixed & 88.00    & 89.04 & 81.42 & 71.23 & 94.19 & 84.78          & 8.83  \\
7 & $\mathbf{FN_{T}}$ & Vis. \& Text & pre-trained & variable & 83.21 & 92.29 & 84.03 & 83.69 & 83.56 & 85.36          & \textbf{3.89}  \\
8 & $\mathbf{FN_{MLP}}$ & Vis. \& Text & pre-trained & variable & 76.38 & 96.1  & 84.03 & 76.87 & 94.85 & 85.65          & 9.48  \\
9 & $\mathbf{FN_{T}}$ & Vis. \& Text & fine-tuned & fixed & 79.12 & 85.04 & \underline{85.25} & 79.88 & \textbf{96.84} & 85.23          & 7.08  \\
10 & $\mathbf{FN_{MLP}}$ & Vis. \& Text & fine-tuned & fixed & \underline{90.75}
 & \underline{96.64}  & 78.76 & 73.97 & 97.13 & 87.45          & 10.56 \\
11 & $\mathbf{FN_{T}}$ & Vis. \& Text & fine-tuned & variable & 85.52 & 92.09 & \textbf{86.50}  & 83.27 & \underline{96.52} & \underline{88.78}          & \underline{5.41}  \\
12 & $\mathbf{FN_{MLP}}$ & Vis. \& Text & fine-tuned & variable & \textbf{96.37} & \textbf{98.60}  & \underline{85.25} & 78.52 & 94.02 & \textbf{90.55} & 8.42 \\ \hline \hline
    \end{tabular}
\end{table*}

\begin{figure}[t!]
\centering
\setlength{\abovecaptionskip}{5pt}
    \includegraphics[width=1\linewidth]{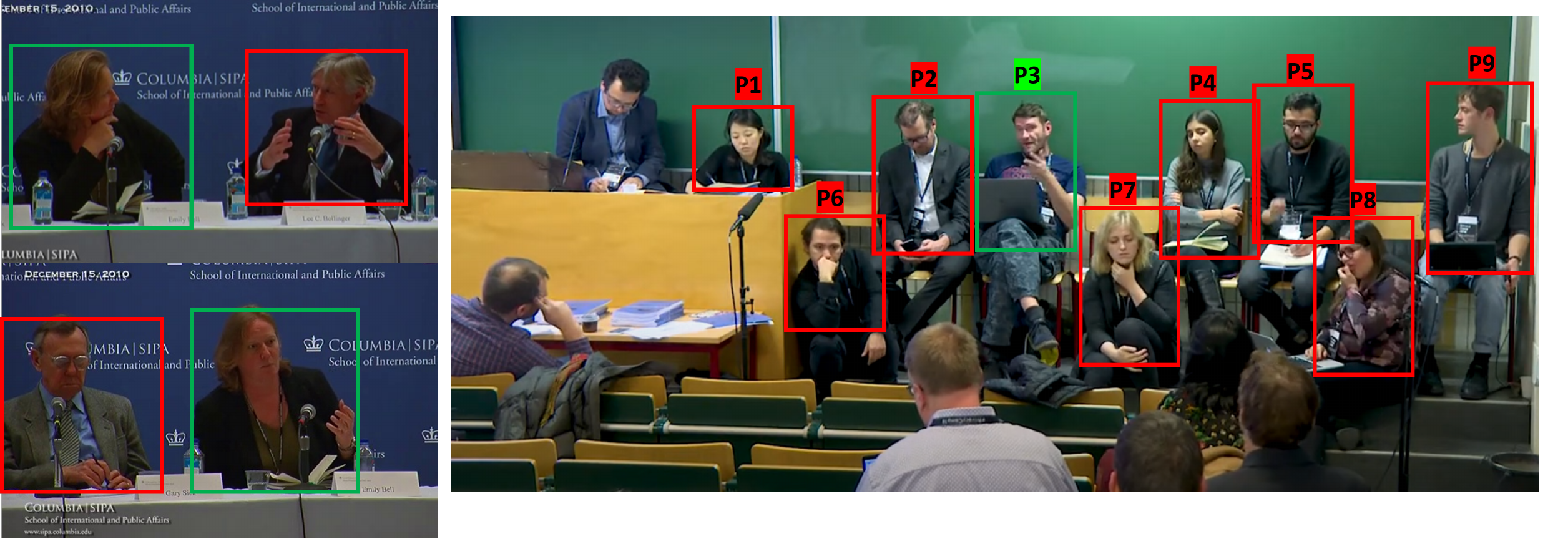}
    \caption{Example frames from the datasets used in this paper. On the left are the Columbia and Modified Columbia datasets, and on the right is the RealVAD dataset. Green boxes indicate the active speakers, while red boxes denote other participants with VAD ground-truth, who are not speaking at that moment.}
    \label{fig:datasets}
    \Description{Datasets used.} 
\end{figure}

\section{Experimental Analysis \& Results}
\label{sec:exp}

The experimental analysis includes comparisons with the SOTA and an ablation study, where different combinations of the CLIP-VAD were tried along with assessing the VLM's performance for VAD. For evaluation purposes, we utilize Columbia \cite{VAD_Audio_Supervision}, Modified Columbia \cite{VAD_Motion_Segmentation}, and RealVAD \cite{RealVAD}, which are three popular benchmarks for visual VAD and particularly were used by the SOTA presenting upper body activity-based VAD (see Fig. \ref{fig:datasets}). 

Columbia dataset \cite{VAD_Audio_Supervision} contains an 87-minute video of a panel discussion including several individuals' speaking activity annotations, in which 2-3 speakers are visible at a time.
The Modified Columbia dataset \cite{VAD_Motion_Segmentation} is derived from the Columbia dataset \cite{VAD_Audio_Supervision}. Unlike Columbia dataset, which contains a significant number of not speaking frames, Modified Columbia has more balanced classes. This more balanced distribution makes the evaluation less skewed, leading to more reliable assessments while the training splits contain fewer samples than Columbia, as noted in \cite{VAD_Motion_Segmentation}. Following SOTA, the evaluations of the Columbia and the Modified Columbia datasets are performed for five panelists: Bell, Bollinger, Lieberman, Long, and Sick. While the Columbia dataset \cite{VAD_Audio_Supervision} includes bounding box annotations for the head position of each panelist, herein, for both Columbia and Modified Columbia, we use the upper body crops supplied by \cite{VAD_UBM,primitive}. 
On the other hand, the RealVAD dataset \cite{RealVAD} comprises an 83-minute panel discussion featuring panelists from various ethnic backgrounds, including British, Dutch, French, German, Italian, American, Mexican, Columbian, and Thai.
The video is recorded using a static, mounted camera, capturing the nine panelists in a full shot. The panelists sit in two rows and are engaged in various activities, leading to potential partial occlusions of the upper body.

The standard evaluation settings of visual VAD, including \textbf{leave-one-person-out} cross-validation (i.e., in each fold of cross-validation, the testing set comprises data from a single individual, while the training set incorporates data from all other individuals) and \textbf{F1-score} as the evaluation metric are used. Results are presented as the F1-score for each person, together with the average and standard deviation across all individuals. The application of leave-one-person-out cross-validation facilitates evaluating the VAD model's ability to generalize to unseen individuals, taking into account the variability in head and body motion patterns among different people, known as the domain-shift problem \cite{RealVAD}.

\begin{table}
    \centering
    \caption{Average and standard deviation of the performances of various fusion networks ($FN$) on the Columbia \cite{VAD_Audio_Supervision} and RealVAD \cite{RealVAD} datasets in terms of F1-score (\%).
    The best results are in black. Exp. indicates the index of the experiments given in Table \ref{tab:ablation}.}
\label{tab:ablation2}

    \begin{tabular}{llll}  \hline \hline
           \textbf{Dataset} &
           \textbf{Exp.} &
           \textbf{Avg.} &
           \textbf{Std.} \\ \hline \hline
Columbia \cite{VAD_Audio_Supervision} & 12 & 93.8 & \textbf{3.7}  \\
Columbia \cite{VAD_Audio_Supervision} & 11 & \textbf{95.2} & 4.9 \\ \hline
RealVAD \cite{RealVAD} & 12 & 86.4  & 6.3  \\
RealVAD \cite{RealVAD} & 11 & \textbf{88.2} & \textbf{5.3} \\ \hline \hline
    \end{tabular}
\end{table}

\begin{figure}[h!]
\centering
\setlength{\abovecaptionskip}{5pt}
    \includegraphics[width=0.95\linewidth]{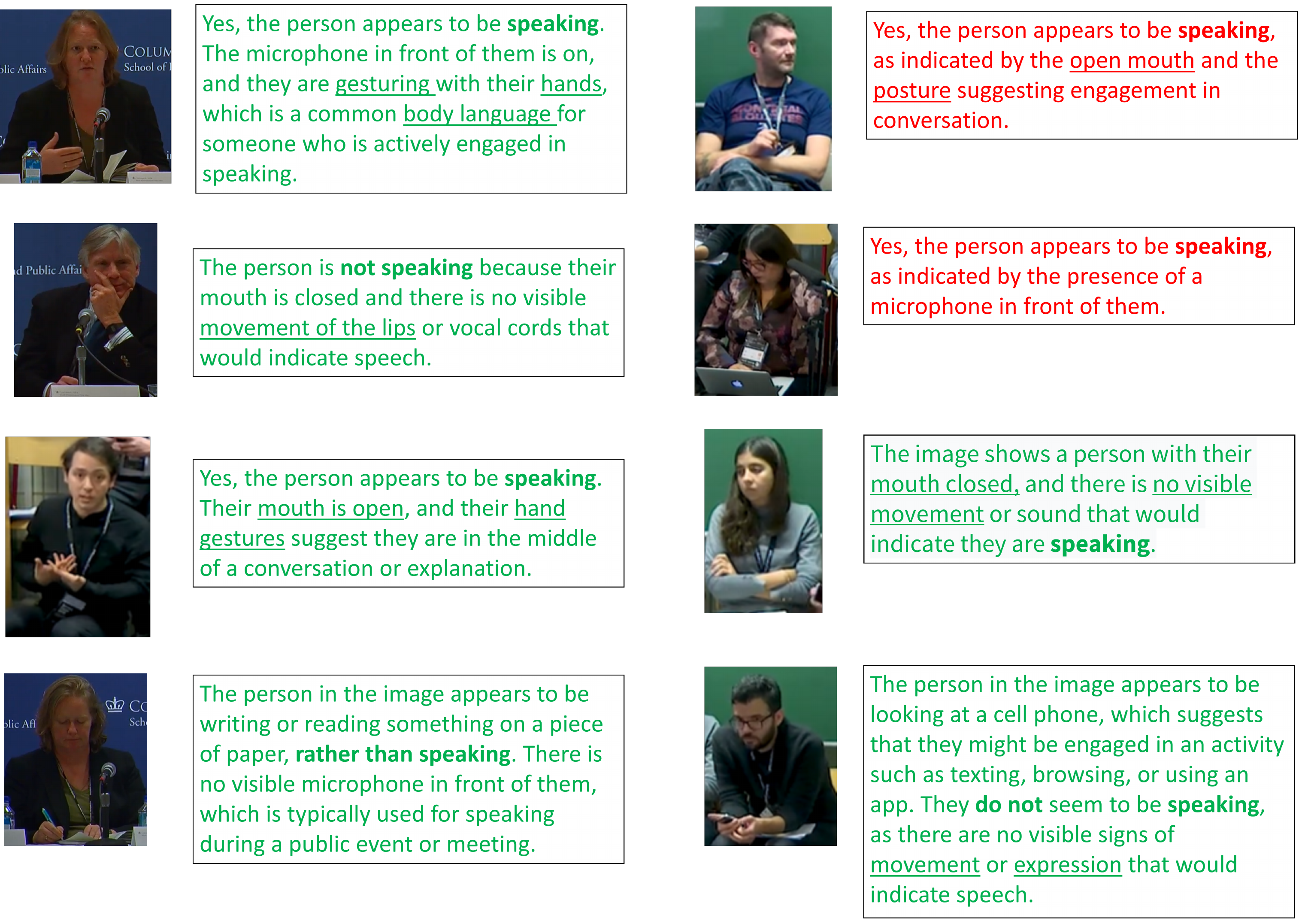}
    \caption{Example text responses obtained upon using LLaVA \cite{Llava_paper1,Llava_paper2} with the second prompt: \textit{Is the person speaking? Explain why in a few words.} The \textcolor{green}{green texts} are the cases where the VAD class is correctly predicted, while the \textcolor{red}{red texts} signify instances of incorrect predictions.}
    \label{fig:llava}
    \Description{LLava output.} 
\end{figure}

\begin{table*}[!ht]
\caption{Comparisons on Columbia dataset \cite{VAD_Audio_Supervision}. We report F1-scores (\%) for each person, the overall average (AVG), and the overall standard deviation (STD).
V, AV, and VT stand for visual, audio-visual, and visual and generated text modalities, respectively. The last two rows show the best of all results excluding CLIP-VAD for audio-visual (denoted as max-AV) and visual (shown as max-V) VAD methods, respectively. Bold results indicate the best of all for each column.
The \colorbox{corn}{colored results} are the ones in which the CLIP-VAD performs better than at least one of the max-AV and max-V.}
\label{tab:Columbia}
\centering
\resizebox{0.8\linewidth}{!}{%
\begin{tabular}{lccccccccc}
\hline \hline
Method & Venue   & Modality     & Bell & Boll & Lieb & Long & Sick & AVG   & STD      \\ \hline \hline

\cite{VAD_Audio_Supervision} & ECCV 2016 & V         & 82.9 & 65.8 & 73.6 & 86.9 & 81.8 & 78.2  & 8.5 \\

SyncNet \cite{chung2017out} & ACCV 2017 & AV & 93.7 & 83.4 & 86.8 & 97.7 & 86.1 & 89.5 & 5.9 \\

\cite{VAD_UBM} & ICCV 2019  & V         & 89.2 & 88.8 & 85.8 & 81.4 & 86.0   & 86.2 & 3.1 \\

RGB-DI \cite{VAD_UBM} & ICCV 2019        & V         & 86.3 & 93.8 & 92.3 & 76.1 & 86.3 & 86.9 & 7.0 \\

LWTNet \cite{afouras2020self} & ECCV 2020 & AV & 92.6 & 82.4 & 88.7 & 94.4 & 95.9 & 90.8  & 5.4 \\

RealVAD \cite{RealVAD} & IEEE TMM 2020  & V         & 92.0   & \textbf{98.9} & 94.1 & 89.1 & 92.8 & 93.4 & 3.6 \\

S-VVAD  \cite{VAD_Motion_Segmentation} & WACV 2021   & V         & 92.4 & 97.2 & 92.3 & 95.5 & 92.5 & 94.0 & \textbf{2.2} \\
\cite{truong2021right} & CVPR 2021   & AV & 95.8 & 88.5 & 91.6 & 96.4 & 97.2 & 93.9  & 3.7 \\
TalkNet \cite{tao2021someone} & ACM MM 2021 & AV & 97.1 & 90.0   & \textbf{99.1} & 96.6 & 98.1 & \textbf{96.2} & 3.6 \\
UNICON  \cite{zhang2021unicon} & ACM MM 2021 & AV & 93.6 &81.3 & 93.8 & 93.5 & 92.1	& 90.9 & 5.4 \\
ACLNet \cite{xiong2022look} & IEEE TMM 2022 & AV & \textbf{97.4} & 88.1 & 97.5 & 98.5 & 98.0   & 95.9  & 4.4 \\
GSCMIA  \cite{sharma2023audio} & IEEE JSTSP 2023 & AV & 96.3 & 89.4 & \textbf{98.7} & 98.7 & 96.8 & 96.0 & 3.8 \\
CLIP-VAD (\textbf{Ours}) &  & VT  & \cellcolor{corn} 96.9 &	86.7 &	\cellcolor{corn} 96.0 & \cellcolor{corn} 97.8	& \cellcolor{corn} \textbf{98.8} & 95.2
 &	4.9
       \\ \hline
      & & max-AV &	98.1 &	89.4 &	99.1 &	99.3 &	98.1 & 	 & \\
      & & max-V	& 92.4 &	98.9 &	94.1 &	95.5 &	92.8 &	 & \\ \hline \hline

\end{tabular}
}
\end{table*}

\begin{table}[!h]
\caption{Comparisons on Modified Columbia dataset \cite{VAD_Motion_Segmentation}
Bold indicates the best. We report F1-scores (\%) for each person, the
overall average (AVG), and the overall standard deviation (STD).}
\label{tab:ModCol}
\centering
\resizebox{\linewidth}{!}{%
\begin{tabular}{lccccccc}
\hline \hline
& Bell & Boll & Lieb & Long & Sick & AVG & STD      \\ \hline  \hline
S-VVAD  \cite{VAD_Motion_Segmentation} & 86.1 & 87.7 & 96.7 & 84.0 & 75.1 & 85.9 & \textbf{7.8} \\
CLIP-VAD (\textbf{Ours}) & \textbf{96.4} & 78.5 & \textbf{94.0} & \textbf{85.2} & \textbf{98.6} & \textbf{90.6} & 8.4 \\ \hline \hline
\end{tabular}
}
\end{table}

\begin{table*}[!ht]
\caption{Comparisons on RealVAD dataset \cite{RealVAD}.
Bold indicates the best. We report F1-scores (\%) for each panelist, the
overall average (AVG), and the overall standard deviation (STD). The same pre-training, and training setups are color-coded for clarity. Zero-shot and fine-tuning experiments are seperated by a double horizontal line.}
\label{tab:RealVAD}
\centering
\resizebox{\linewidth}{!}{%
\begin{tabular}{l|ccccccccccccccc}
\hline \hline

Method        & Modality  & Pre-training Data & Training Data & Testing Data   & P1   & P2   & P3   & P4   & P5   & P6   & P7   & P8   & P9   & AVG  & STD \\ \hline

\rowcolor{LightOrange}
\cite{RealVAD} & V & Columbia \cite{VAD_Audio_Supervision} & - & RealVAD \cite{RealVAD}	& 53.6 & 51.1 &	41.1 &	50.2 & 37.3 & 50.3 &	56.8 &	53.6 &	69.8 &	51.5 &	9.3 \\ 

\rowcolor{LightPink}
UNICON  \cite{zhang2021unicon}       & V & AVA-ActiveSpeaker \cite{roth2020ava} & - & RealVAD \cite{RealVAD}  & 86.7 & 78.1 & 70.5 & 73.1 & 68.9 & 84.9 & 93.0 & 80.4 & 87.0 & 80.3 & 7.8 \\

\rowcolor{LightPink}
UNICON  \cite{zhang2021unicon}  & AV & AVA-ActiveSpeaker \cite{roth2020ava} & - & RealVAD \cite{RealVAD}  & \textbf{94.3} & 74.0 & \textbf{89.9} & 76.7 & \textbf{80.6} & \textbf{93.6} & \textbf{98.8} & 83.5 & \textbf{93.5} & \textbf{87.2} & 8.3 \\ 

\rowcolor{LightOrange}
CLIP-VAD (\textbf{Ours}) & VT & Columbia \cite{VAD_Audio_Supervision} & - & RealVAD \cite{RealVAD}	& 89.0 &	\textbf{81.6}	& 81.4 &	\textbf{83.4} &	79.3 &	\textbf{93.6} &	97.2 &	\textbf{85.8} &	93.4 &	\textbf{87.2} &	\textbf{6.4} \\\hline \hline


\rowcolor{LightCyan}
\cite{RealVAD} & V & - & RealVAD \cite{RealVAD} & RealVAD \cite{RealVAD}   & 51.6 & 53.5 & 42.9 & 51.7 & 44.4 & 50.5 & 58.7 & 67.9 & 55.8 & 53.0 & 7.1 \\

\rowcolor{LightGreen}
UNICON  \cite{zhang2021unicon}        & V & AVA-ActiveSpeaker \cite{roth2020ava} & RealVAD \cite{RealVAD} & RealVAD \cite{RealVAD}  & 86.9 & 76.5 & 81.6 & \textbf{87.0} & 79.6 & \textbf{88.9} & 97.0 & 84.5 & 88.9 & 85.6 & 5.7 \\ 

\rowcolor{LightGreen}
UNICON  \cite{zhang2021unicon}        & AV & AVA-ActiveSpeaker \cite{roth2020ava} & RealVAD \cite{RealVAD} & RealVAD \cite{RealVAD}  & \textbf{96.5} & \textbf{81.1} & \textbf{86.9} & 84.4 & \textbf{89.9} & 85.6 & 94.9 & 88.1 & \textbf{90.9} & \textbf{88.7} & \textbf{4.7} \\

\rowcolor{LightCyan}
CLIP-VAD (\textbf{Ours}) & VT & - & RealVAD \cite{RealVAD} & RealVAD \cite{RealVAD}   & 91.7	& 78.8 &	86.2 &	87.7& 	84.4 &	87.8 &	\textbf{98.5} &	\textbf{88.9} &	89.5 &	88.2 &	5.3 \\
 \hline
\hline
\end{tabular}
}
\end{table*}

\subsection{Ablation Study}
\label{sec:ablation}
The results of the ablation study, in which the different combinations of the CLIP-VAD are tested, are given in Tables \ref{tab:ablation} and \ref{tab:ablation2}. Below, we delve into a detailed discussion of our findings.

As mentioned in Sec. \ref{sec:ImpDet}, we use the two sets of prompts. In the case of the first prompt, the responses consistently result in either ``yes'' or ``no''. In the second prompt, we aim to test the expressive capabilities of the model, obtaining more complex and varied responses such as \textit{The person is not speaking because their mouth is closed and there is no visible movement of the lips.} (See Fig. \ref{fig:llava} for additional examples obtained through the use of LLaVA.).
Using \emph{prompt 1} allows us to evaluate the standalone LLaVA-13B's \cite{Llava_paper1,Llava_paper2} VAD performance (Exp. 1 of Table \ref{tab:ablation}). Such an analysis leads to overall the worst performance, particularly for subjects Sick and Long where the F1-scores are below $60\%$. 

Furthermore, the textual responses produced by LLaVA-13B were converted into textual embeddings utilizing the pre-trained CLIP text encoder with ResNet-101. Analysis of the cosine similarity between these embeddings revealed that, on average, the captions generated by the first prompt exhibited a similarity exceeding 0.9. As a result, it can be inferred that classifiers, which would later utilize these embeddings as input, might struggle to discern significant differences between descriptions of a person speaking and not speaking. 
Therefore, we also tested to convert the results of the first prompt such that instead of ``yes'', we used \textit{the person is engaged in a conversation} and instead of ``no'', we used \textit{no one is talking} captions, which are referred as ``fixed'' in Table \ref{tab:ablation}. As a consequence of this change, the cosine similarity was found to be 0.75, which was assumed to have substantial differences compared to others when used, e.g., for multimodal classification. 
To sum up, ``yes/no'' captions were only used to evaluate the classification performance of LLaVA-13B. The highly variable and expressive captions generated by the second prompt were used as textual inputs for the multimodal models referred to as \textit{variable} in Table \ref{tab:ablation}. Additionally, the \textit{fixed} captions: \textit{the person is engaged in a conversation} and \textit{no one is talking} were used as alternative textual inputs for the multimodal models.

To test the performance of the pre-trained CLIP visual encoders ResNet101 and ViT-B/16 on the VAD task, we attach an MLP whose design is as described in Sec. \ref{sec:ImpDet} since the visual encoders alone lack classification capability.
From the observed results (Exp. 2 \& 3 of Table \ref{tab:ablation}), it became evident that the performance of ResNet101 not only competes with the more complex ViT-B/16 but is even superior, with an average F1-score greater by 8\%. Furthermore, it can be observed how the domain-shift problem has influenced the VAD results, particularly for Bell and Long, for both visual encoders, leading to relatively lower performances with respect to the others. 

We further tried to test also the fine-tuned ResNet101 (Exp. 4) for the VAD task considering its better performance than ViT-B/16.
Such results indicate a clear improvement over using the pre-trained CLIP for almost all individuals. In particular, there is an average improvement in the F1-score of $\sim$3\% compared to the score obtained by classifying embeddings from pre-trained ResNet-101. In particular, from these results, it is observed that the domain-shift problem has been effectively mitigated for many panelists compared to the pre-trained visual encoder. It is also noticeable that the performance of Bollinger and Lieberman has slightly worsened.

The other experiments involve various combinations of CLIP-VAD, wherein the CLIP encoders remain fixed to ResNet101 while the $FN$ changes across MLP or transformer, and the visual encoder is utilized as pre-trained or fine-tuned, with text embeddings fixed or variable (Exp. 5-12). 
Despite the $FN_{MLP}$ with variable captions and fine-tuned visual encoder emerging as the best-performing model on average, there are instances where $FN_{T}$ outperforms $FN_{MLP}$ (e.g., in Exp. 5 and 6, for Lie., Bell and on average). Overall, fine-tuning and employing variable captions enhance performance across all individuals as well as on average.

We additionally present the results of $FN_{MLP}$ and $FN_{T}$, employing variable captions and fine-tuned visual encoders, across both the Columbia \cite{VAD_Audio_Supervision} and RealVAD \cite{RealVAD} datasets in Table \ref{tab:ablation2}. Notably, for these datasets where larger training sets per fold are available, one can observe the superior performance of $FN_{T}$ over $FN_{MLP}$.

\subsection{Comparisons with the SOTA}
\label{sec:compSOTA}
We compare the effectiveness of CLIP-VAD on Tables \ref{tab:Columbia}, \ref{tab:ModCol}, and \ref{tab:RealVAD} for Columbia \cite{VAD_Audio_Supervision}, Modified Columbia \cite{VAD_Motion_Segmentation} and RealVAD \cite{RealVAD} datasets, respectively.
Overall, CLIP-VAD outperforms all visual VAD approaches as well as some of the audio-visual methods. Considering that the used benchmarks include a single speaker at a time, without overlapping speech, we claim that audio signal can help significantly to detect whether there is a speaker or not at a certain time. While our focus is not to outperform all audio-visual VAD methods, recognizing the crucial role of audio signals in this task, we find CLIP-VAD's performance in surpassing certain audio-visual models noteworthy. Moreover, such results supports the potential to integrate CLIP-VAD with audio signals to enhance overall performance.
In detail, for Columbia dataset \cite{VAD_Audio_Supervision} (Table \ref{tab:Columbia}), CLIP-VAD surpasses the performances of all the visual VAD methods as well as audio-visual methods: SyncNet \cite{chung2017out}, LWTNet \cite{afouras2020self}, \cite{truong2021right}, and UNICON \cite{zhang2021unicon}, on average.
It is important to notice that the performance of CLIP-VAD on panelist Sick is the best of all methods. Furthermore, when we examine the best performance achieved among the visual VAD methods (indicated as max-V in the Table \ref{tab:Columbia}) and among the audio-visual methods (denoted as max-AV in the Table \ref{tab:Columbia}), it becomes apparent that in four out of five cases, CLIP-VAD outperforms at least one of the two. Consequently, it can be argued that the visual backbones (e.g., ResNet50) utilized in such methods can be substituted with CLIP visual encoder through fine-tuning, and furthermore, the text descriptions encoded with CLIP text encoder can contribute to enhancing performance.

For the Modified Columbia dataset \cite{VAD_Motion_Segmentation} (Table \ref{tab:ModCol}), CLIP-VAD outperforms S-VVAD \cite{VAD_Motion_Segmentation} on average by 5\% in terms of F1-score. CLIP-VAD only falls short of the S-VVAD method \cite{VAD_Motion_Segmentation} for Boll. It is worth noting that the Modified Columbia dataset is more challenging than the Columbia dataset, as it has less training data. Despite this, CLIP-VAD still manages to outperform the SOTA. 

For the RealVAD dataset, there is no consistent evaluation approach across the literature. In this study, we adopt the evaluation approach utilized in the original paper \cite{RealVAD}, which comprises two methods: the first involves traditional training and testing on the same dataset, while the second entails training on the Columbia dataset \cite{VAD_Audio_Supervision} and subsequently testing the trained model on RealVAD \cite{RealVAD}. This latter approach is referred to as zero-shot and/or cross-dataset evaluation. On the other hand, Zhang et al. \cite{zhang2021unicon} conducted their evaluation on this dataset by initially pre-training their model on a large dataset known as AVA-ActiveSpeaker \cite{roth2020ava}. They then either applied the trained model directly to RealVAD (hence zero-shot) or fine-tuned the model on RealVAD before testing it on the same dataset. Given the scale of AVA-ActiveSpeaker \cite{roth2020ava} and the training of UNICON \cite{zhang2021unicon} on both audio and visual modalities, especially in the fine-tuned scenario, UNICON holds a significant advantage over CLIP-VAD and the approach by Beyan et al. \cite{RealVAD}, as it has access to a larger and more diverse dataset. Additionally, considering that the RealVAD dataset \cite{RealVAD} typically features single speakers at a time (with no multiple speakers or overlapped speech), audio-visual methods like UNICON \cite{zhang2021unicon} may have an edge over video-only VAD methods in accurately detecting whether anybody is speaking or not using the audio signal. 
Nevertheless, as seen in Table \ref{tab:RealVAD}, CLIP-VAD achieves an on-par performance with the audio-visual UNICON \cite{zhang2021unicon} in the zero-shot setting on average (87.2\% F1-score), while it clearly outperforms the visual UNICON \cite{zhang2021unicon} (87.2\% versus 80.3\% F1-score) and RealVAD \cite{RealVAD} (87.2\% versus 51.5\% F1-score) in both average performance and across all panelists. On the other hand, in fine-tuning experiments, audio-visual UNICON surpasses CLIP-VAD on average by 0.5\% F1-score and for several panelists (i.e., P1, P2, P3, and P5), which we attribute to its pre-training on a large multimodal dataset. Conversely, CLIP-VAD outperforms visual-only UNICON \cite{zhang2021unicon} with +2.6\% and RealVAD \cite{RealVAD} with +35.2\%, on average.

\section{Discussions and Conclusions}
\label{sec:conc}

We have introduced a Voice Activity Detection (VAD) method employing Vision-Language Models (VLMs). It takes as input a video segment composed of a single person's upper body to predict their VAD label as ``speaking'' or ``not speaking''. 
While this video segment is input to the visual encoder of a certain VLM, called CLIP \cite{CLIP}, the central frame of the video segment is used to automatically generate the text descriptions regarding the speaking activity of the person by VLM model LLaVA-13B \cite{Llava_paper1, Llava_paper2} through prompt engineering, which is further given as the input to the text encoder of CLIP \cite{CLIP}. The visual and textual encodings in the latest stage are fused with a simple concatenation and learned with a deep model to perform VAD predictions.

This study is a pioneer in showcasing the contribution of visual-textual models, which improves VAD results, particularly compared to VAD methods relying solely on visual data. It also presents the VAD performance of a standalone VLM model using a single image and a rather simple prompt. Even though the standalone VLM is not sufficient by itself, the text descriptions generated by it contribute to achieving improved VAD when combined with visual features. This highlights the effectiveness of the video-language coupling, as demonstrated across various downstream tasks, suggesting its efficacy over relying solely on visual cues for VAD. 
Furthermore, we empirically demonstrate that the proposed method can perform comparably or even better than several audio-visual VAD approaches. This accomplishment is significant, particularly considering that audio is a primary modality for VAD tasks, as indicated by prior studies. Having the audio in the loop is especially effective in scenarios where there are no simultaneous speakers, as observed in the benchmarks utilized in this study.
Additionally, the noteworthy performance of our CLIP-VAD is evident, especially considering that some SOTA models have been pre-trained on large audio-visual datasets.
Such conclusions about the performance of CLIP-VAD are not only applicable to the traditional setting, where the training and testing data share the same distribution (i.e., come from the same dataset), but also to zero-shot setting in which the model is trained on one dataset and tested on another dataset without fine-tuning.
\\

\noindent \textbf{{Limitations \& Future Work.}}
The standard deviation results of CLIP-VAD are not satisfactory compared to other methods. This shows that CLIP-VAD is not resilient to the domain-shift problem occurring across different speakers whose head and body activity can be more evident while some others use less head and body expression when they are speaking. Indeed, our method neither integrates a domain adaptation procedure such as applied in \cite{VAD_UBM} nor is pre-trained on a large VAD dataset as applied in \cite{zhang2021unicon} which potentially can help to generalize better across different individuals.

Moreover, we exclusively assessed LLaVA-13B \cite{Llava_paper1, Llava_paper2} both as a standalone VLM and integrated within the CLIP-VAD framework. While this choice can be justified given LLaVA's demonstrated effectiveness in earlier works, it is worth noting that there are several other models that warrant exploration to provide a more comprehensive evaluation. Besides, we have not aimed to test an ensemble of prompts that can potentially improve the results. 

Despite our approach to handling temporal data in the visual domain, we extract textual embeddings from single images. This decision was driven by the need to optimize model efficiency, as obtaining captions for each video frame can be resource-intensive. However, we presume that the use of 10 video frames would not significantly increase the variance in captions. Nonetheless, exploring the processing of spatio-temporal data with VLMs could potentially enhance performance, although this remains an area of investigation beyond VAD specifically.

The obtained results suggest the potential benefits of integrating VLM-based pipelines into audio-visual VAD systems, which could enhance performance. Future work will further explore this topic in depth.

\bibliographystyle{ACM-Reference-Format}
\bibliography{main}

\end{document}